\documentclass{article}

    \usepackage[preprint]{tackling_climate_workshop_style}

\usepackage{tikz}
\usetikzlibrary{shapes.geometric, arrows.meta, positioning}
\usepackage{amsmath}
\usepackage[utf8]{inputenc} 
\usepackage[T1]{fontenc}    
\usepackage{hyperref}       
\usepackage{url}            
\usepackage{booktabs}       
\usepackage{amsfonts}       
\usepackage{nicefrac}       
\usepackage{microtype}      
\usepackage{graphicx}       
\usepackage{float}          
\usepackage{subcaption}     
\usepackage{svg}            
\usepackage{booktabs}       
\usepackage{multirow}

\title{DWaste: Greener AI for Waste Sorting using Mobile and Edge Devices}

\author{%
  Suman Kunwar \\
  DWaste, USA\\
  \texttt{sumn2u@gmail.com} \\
}

\begin{document}

\maketitle

\begin{abstract}

The rise of convenience packaging has led to generation of enormous waste, making efficient waste sorting crucial for sustainable waste management. To address this, we developed DWaste, a computer vision-powered platform designed for real-time waste sorting on resource-constrained smartphones and edge devices, including offline functionality. We benchmarked various image classification models (EfficientNetV2S/M, ResNet50/101, MobileNet) and object detection (YOLOv8n, YOLOv11n) including our purposed YOLOv8n-CBAM model using our annotated dataset designed for recycling. We found a clear trade-off between accuracy and resource consumption: the best classifier, EfficientNetV2S, achieved high accuracy ($\approx 96\%$) but suffered from high latency ($\approx 0.22\text{s}$) and elevated carbon emissions. In contrast, lightweight object detection models delivered strong performance (up to $80\%\text{ mAP}$) with ultra-fast inference ($\approx 0.03\text{s}$) and significantly smaller model sizes ($<7\text{MB}$), making them ideal for real-time, low-power use. Model quantization further maximized efficiency, substantially reducing model size and VRAM usage by up to $75\%$. Our work demonstrates the successful implementation of "Greener AI" models to support real-time, sustainable waste sorting on edge devices.

\end{abstract}

\keywords{Model Quantization, Edge Computing, Object Detection, Waste Management, Greener AI}

\section{Introduction}

The growth of convenience packaging has increased waste generation \cite{pinos_why_2022}, underscoring the need for efficient sorting. Global waste is projected to grow from 2.1 to 3.8 billion tons by 2050 \cite{unep_beyond_2024}, an increase that compounds the financial, environmental and planetary burdens. For instance, a study on Chilean municipal solid waste (MSW) found the cost of the unsorted waste to be 297.66 euro per ton \cite{sala-garrido_monetary_2023}. Furthermore, waste contamination poses a significant challenge to implement a circular economy, particularly given that the US has seen its recycling rate stagnate at around 35\% for over a decade \cite{turner_problems_2024}.

To address this challenge, traditional waste management has begun incorporating technology. Over the past decades, various machine learning (ML) models such as linear regression (LR), support vector machine (SVM), and random forest (RF) have evolved for predicting inbound contamination rates \cite{runsewe_machine_2023}. Simultaneously, IoT devices integrated with bins, vehicles, and recycling facilities are aiding waste sorting and data collection, leveraging GPS for route tracking and temperature sensors for fire protection. The recent shift from traditional ML to Deep Learning (DL) has delivered substantial improvements through better computation power and advanced algorithms. However, research remains uneven across sectors, often relying on simplified or artificial data \cite{lu_computer_2022}. 

The economic viability of these systems is a crucial factor, as research by Liu et al. demonstrated that computer vision-enabled systems (CVAS) become cost-effective when labor costs are high, while conventional sorting (CS) is preferable when machinery or maintenance costs are higher; their comprehensive cost model included labor, training, machinery, maintenance, and net present values of investments \cite{liu_computer_2025}. Special DL architectures, particular object detection models, have shown promising results in sorting applications. For example, YOLOv5 models equipped with webcam and robotic arms have demonstrated waste sorting capabilities with 93.3\% accuracy \cite{lahoti_multi-class_2024}. More recently, a YOLOv8 model embedded with a Raspberry Pi achieved 98\% accuracy in complex tasks of real-time intelligent garbage monitoring and collection systems \cite{abo-zahhad_real_2025} showcasing practical, low-cost solutions for smart waste management in urban environments. 

In the realm of classification, high accuracy has been achieved with CNN architectures Ahmad et al. utilized ResNet-based CNN to automatically class 12 waste types, achieving 98.16\% \cite{ahmad_intelligent_2025}. Tran et al. achieved 96\% accuracy using ResNet-50 of organic and inorganic waste classification with raspberry pie 4 to direct sorting \cite{khai_tran_deep_2023}; and a comparison by Soni et al. study, found MobileNet despite achieving 80\% accuracy, offered a superior accuracy and lower computation cost making it highly suitable for scalable real-world applications \cite{soni_mobilenet-based_2024}.

System efficiency, particularly for mobile and edge computing, remains a key consideration, with YOLOv11n proving to be the most power-efficient 125{,}000~$\mu$Ah in 590 seconds), while YOLOv11m/11s performed best in accuracy-driven applications \cite{eva_urankar_waste_2025}. Despite these advancements, challenges persist, as highlighted in a review by Gelar et al. on YOLO and IoT applications, which identified issues with accurate detection, environmental adaptability, and optimizing low-power IoT performance \cite{gelar_systematic_2025}. The Convolutional Block Attention Module (CBAM) integrated with YOLO architectures  to boost feature extraction and spatial attention has shown promising results \cite{xu_improved_2024}.

While our own past study focused on benchmarking models based on accuracy and carbon emission to determine a "greener" classification model, it was limited to classification tasks only \cite{kunwar_managing_2024}. This paper aims to address these challenges by systematically addressing the trade-off in deploying advanced DL models. We benchmark state of the art classification and object detection models including our own model that uses CBAM inhanced backbone and use model quantization as the primary optimization technique, precisely benchmarking the reduction in VRAM usage, model size, and carbon emission to validate the path toward a "Greener AI" solution for waste sorting. Later, the greener model is deployed to mobile apps and edge devices.

\section{Materials and Methods}
The section discusses the dataset used in this study and the methods used to benchmark train the models.

\subsection{Dataset and Preprocessing}
\begin{figure}[H]
  \centering
  \includegraphics[width=0.95\textwidth, height=0.34\textwidth, keepaspectratio]{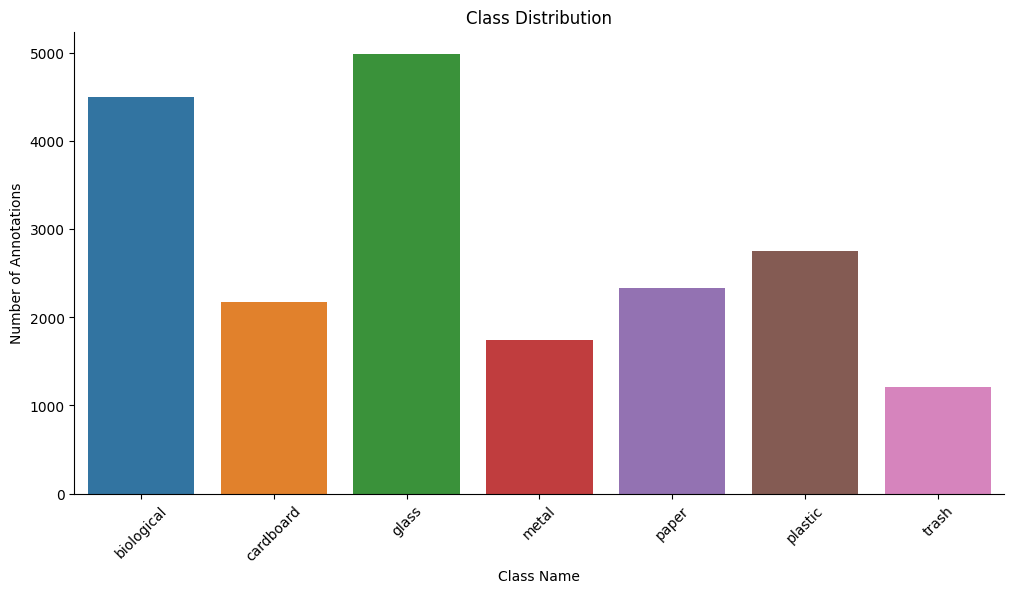}
  \caption{Dataset class distribution}
  \label{fig:image_distribution}
\end{figure}

In this study, we used our garbage dataset \cite{suman_kunwar_dwaste-data-v4-annotated_2025} focusing on seven categories deemed critical for recycling efficiency: biological, cardboard, glass, metal, paper, plastic, and trash. These images were collected from the internet, DWaste platform, and community submissions. All images were annotated with category labels and bounding boxes using Annotate Lab \cite{kunwar_annotate-lab:_2024}. The final processed dataset consists of 11,163 images and 19,700 bounding box instances shown below Figure \ref{fig:image_distribution}.

\begin{figure}[H]
  \centering
  \includegraphics[width=0.95\textwidth]{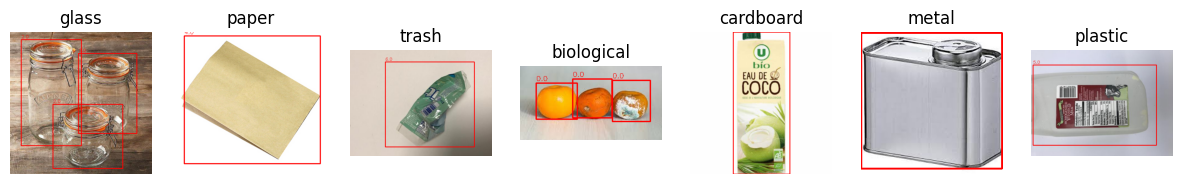}
  \caption{Sample annotated images from our dataset}
  \label{fig:sample_annotated_image}
\end{figure}

The above images show a non-uniform distribution characteristic of real world municipal solid waste streams. For classification models, class imbalance was addressed using undersampling technique, where images were selectively removed from oversampled classes \cite{he_imbalanced_2013}. Conversely, for object detection models, the imbalance was addressed by applying computed class weights during the training phase \cite{singh_review_2023}, which up-weighted the loss contribution from underrepresented classes. The finalized dataset was then partitioned using an 80/20 split for the training and validation sets. The sample annotated waste image of the above classes is shown in Figure \ref{fig:sample_annotated_image}.

\subsection{Model Training}
\begin{figure}[h]
\centering
\resizebox{0.5\columnwidth}{!}{%
\begin{tikzpicture}[node distance=0.5cm and 0.5cm, every node/.style={align=center, font=\Large}]
\tikzstyle{block} = [rectangle, rounded corners, minimum width=3cm, minimum height=0.2cm, draw=black, fill=blue!10, very thick]
\tikzstyle{arrow} = [thick, ->, >=Stealth]

\node (input) [block] {Input\\Image};
\node (backbone) [block, right=of input] {YOLOv8n\\Backbone};

\node (cbam) [block, below=0.6cm of backbone] {CBAM\\(Channel + Spatial Attention)};

\node (head) [block, below=0.6cm of cbam] {Detection\\Head};
\node (output) [block, right=of head] {Output};

\draw [arrow] (input) -- (backbone);
\draw [arrow] (backbone) -- (cbam);
\draw [arrow] (cbam) -- (head);
\draw [arrow] (head) -- (output);

\draw [dashed, gray, thick] (backbone.south) -- (cbam.north);
\draw [dashed, gray, thick] (cbam.south) -- (head.north);

\end{tikzpicture}%
}
\caption{Architecture of the proposed Improved YOLOv8n model with CBAM-enhanced backbone}
\label{fig:improved_yolov8n}
\end{figure}
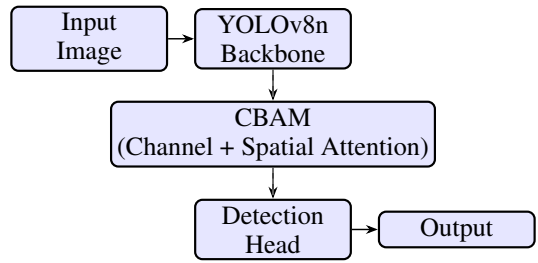

We evaluated both classification (EfficientNetV2S/M, MobileNet, ResNet50/101) and object detection (YOLOv8n, YOLOv11n) architectures including our proposed YOLOv8n-CBAM model shown in Figure \ref{fig:improved_yolov8n}, using a transfer learning approach.

All classification models were initialized with weight pre-trained on the ImageNet dataset. All models were trained for 20 epochs using NVIDIA Tesla T4x2 GPU in Kaggle. The performance of each model was benchmarked using standard performance metrics including Accuracy, Precision, Recall, F1-score, and Mean Average Precision (mAP). With our focus on sustainable AI deployment, we conducted detailed analysis of VRAM usage, model size, and carbon emissions across various phases of the workflow. Carbon emissions were monitored using CodeCarbon library \cite{benoit_courty_mlco2/codecarbon:_2024}. The resulting full-precision models were further optimized via quantization, a technique known to achieve up to 95\% reduction in the number of parameters and model size \cite{francy_edge_2024}, thereby significantly lowering energy use.

\section{Results and Discussion}
Our experiment revealed a distinct tradeoff between model accuracy, size, and carbon efficiency as shown in Table \ref{tab:experimental_results}.
\begin{table}[H]
    \centering
    \renewcommand{\arraystretch}{1.2} 
    \setlength{\tabcolsep}{6pt}        
    \footnotesize                      

    \caption{Experimental Results (Performance metrics and Model Sizes).}
    \label{tab:experimental_results}

    \begin{tabular}{@{}l c c c c c c c@{}}
        \toprule
        \textbf{Model (Classification)} & \textbf{Acc} (\%) & \textbf{P} & \textbf{R} & \textbf{F1} & \textbf{Size (MB)} & \textbf{Q-Size (MB)} \\
        \midrule
        MobileNet                       & 67.50             & 0.67       & 0.68       & 0.67        & \textbf{14.7}               & \textbf{3.5} \\
        EffNetV2M                       & 94.70             & 0.94       & 0.95       & 0.95        & 216.0              & 56.4 \\
        EffNetV2S                       & \textbf{96.00}    & \textbf{0.96} & \textbf{0.96} & \textbf{0.96} & 84.3               & 22.1 \\
        ResNet101                       & 92.10             & 0.91       & 0.93       & 0.92        & 174.6              & 43.6 \\
        ResNet50                        & 91.40             & 0.90       & 0.92       & 0.91        & 97.9               & 24.2 \\
        \midrule
        \multicolumn{7}{@{}l@{}}{\textbf{Metrics:} Acc = Accuracy, P = Precision, R = Recall, F1 = F1 Score.} \\
    \end{tabular}

    \vspace{1.0em} 

    \begin{tabular}{@{}l c c c c c c c@{}}
        \toprule
        \textbf{Model (Detection)} & \textbf{Acc} (\%) & \textbf{P} & \textbf{R} & \textbf{F1} & \textbf{mAP} & \textbf{Size (MB)} & \textbf{Q-Size (MB)} \\
        \midrule
      YOLOv8n & {-} & \textbf{0.78} & 0.65 & 0.75 & 0.76 & 6.5 & 3.1 \\
        YOLOv11n & {-} & 0.77 & 0.69 & 0.77 & 0.77 & \textbf{5.4} & \textbf{2.8} \\
        YOLOv8n-CBAM & {-} & \textbf{0.78} & \textbf{0.73} & \textbf{0.80} & \textbf{0.80} & 6.1 & 3.5 \\
        \bottomrule
        \multicolumn{8}{@{}l@{}}{\textbf{Metrics:} mAP = mean Average Precision. \textbf{Q-Size} = Quantized Size. Best results are in bold.} \\
    \end{tabular}
\end{table}

\begin{figure}[H]
  \centering
  \includegraphics[width=\columnwidth]{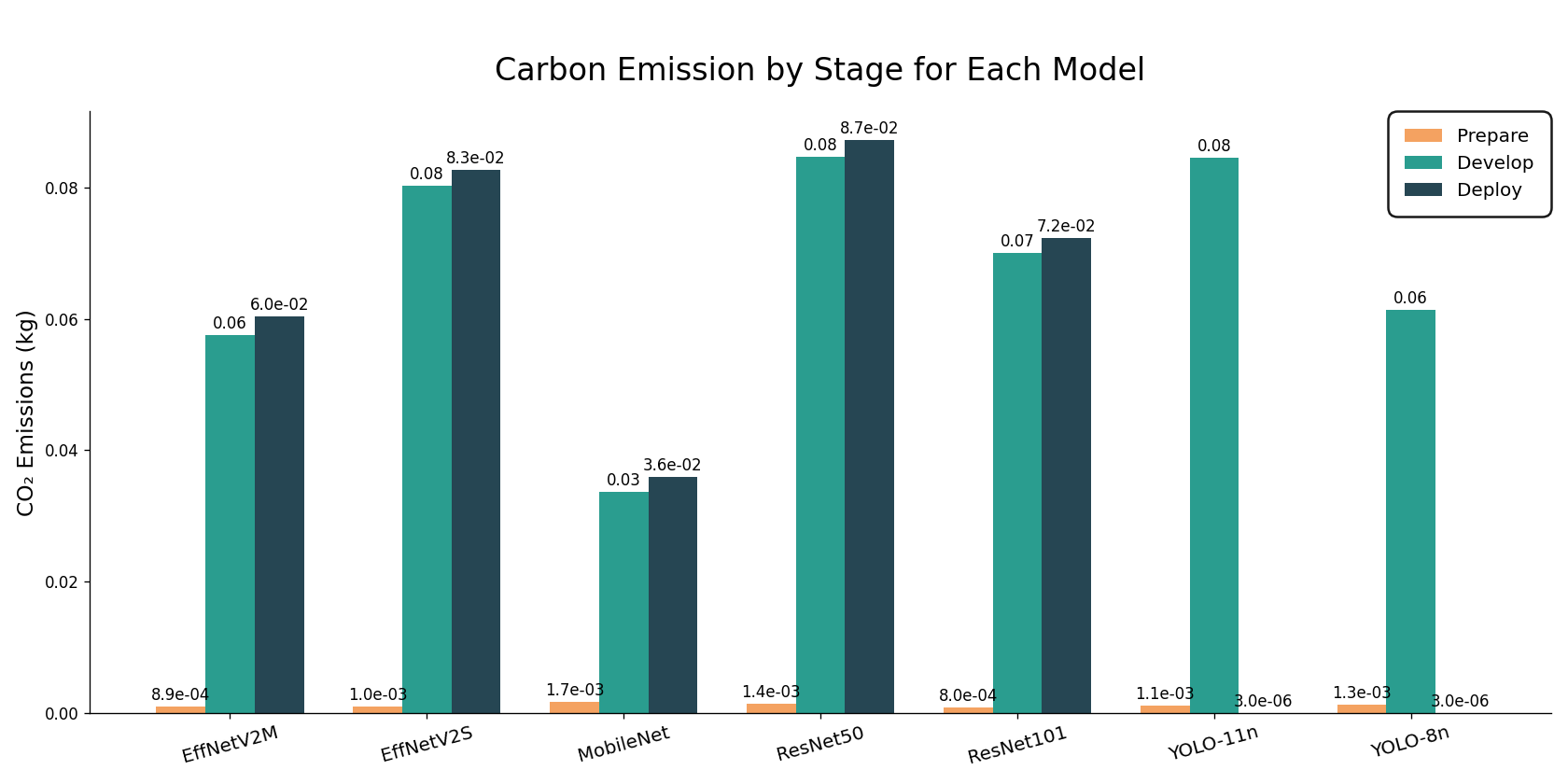}
  \caption{Carbon emission by stage for each model}
  \label{fig:carbon_emission_each_model}
\end{figure}

The high-performance classification models EfficientNetV2M and EfficientNetV2S showed highest accuracy ($\approx 95-96\%$) but was constrained by larger model sizes (216MB and 84.3MB respectively) and consequently resulted in higher training and deployment emissions shown in Figure \ref{fig:carbon_emission_each_model}. Similarly, ResNet101 and ResNet50 delivered strong accuracy (91-92\%) but were also penalized by substantial size and higher emissions. 
In contrast, MobileNet prioritized resource efficiency, exhibiting the smallest initial size (14.7 MB, reduced to 3.5 MB after quantization) with minimal energy usage and lowest accuracy ($\approx 67\%$). The lightweight object detection models, YOLOv8n-CBAM and YOLOv11n demonstrated the most balanced trade-off, achieving $78\%$ precision with impressive quantized sizes ( 
$<3.6,\text{MB}$).

\begin{figure}[H]
  \centering
  \includegraphics[width=\columnwidth]{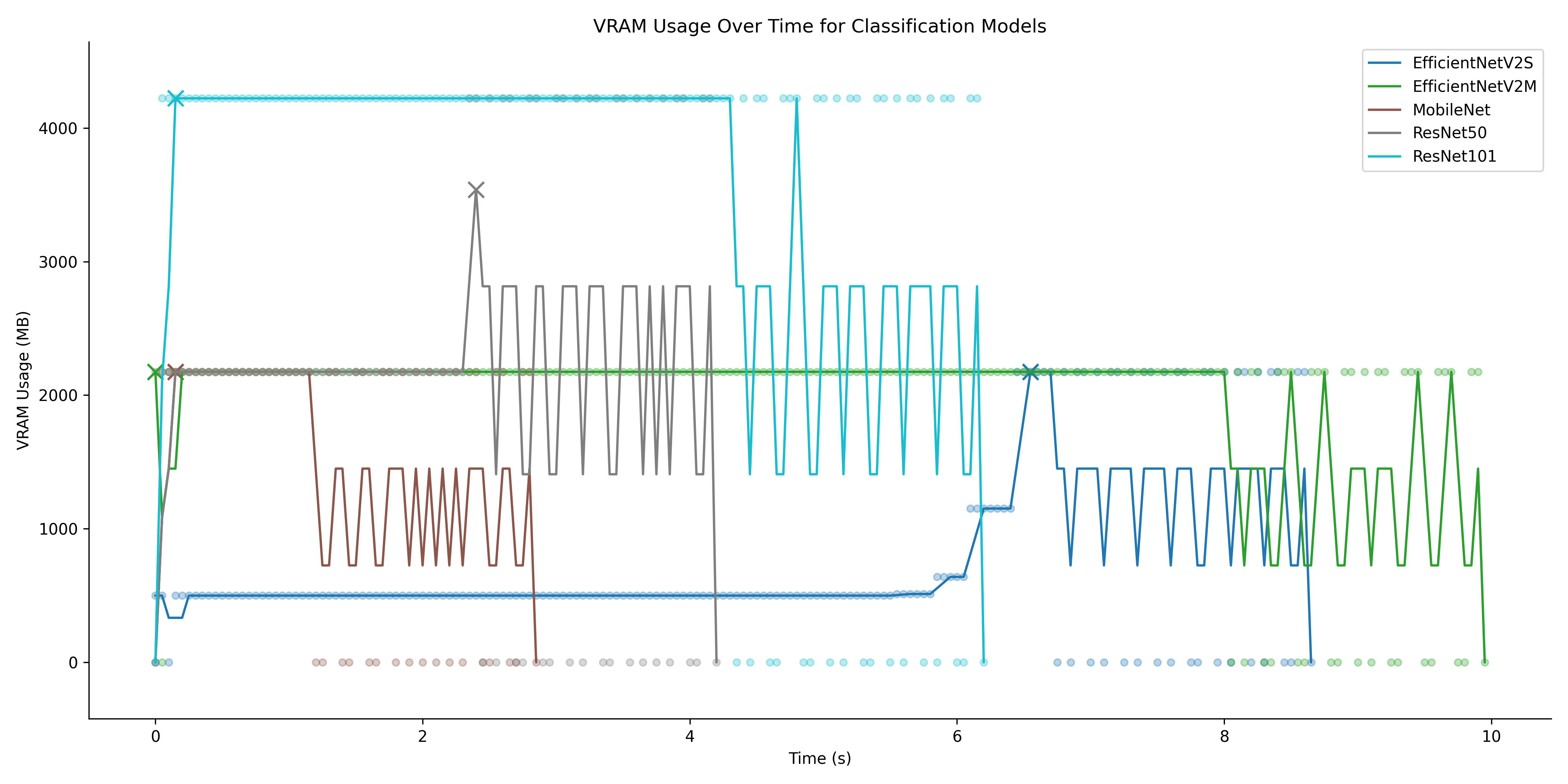}
  \caption{Classification models VRAM usage under inference}
  \label{fig:classification_model_vram_usage}
\end{figure}

\begin{figure}[H]
  \centering
  \includegraphics[width=\columnwidth]{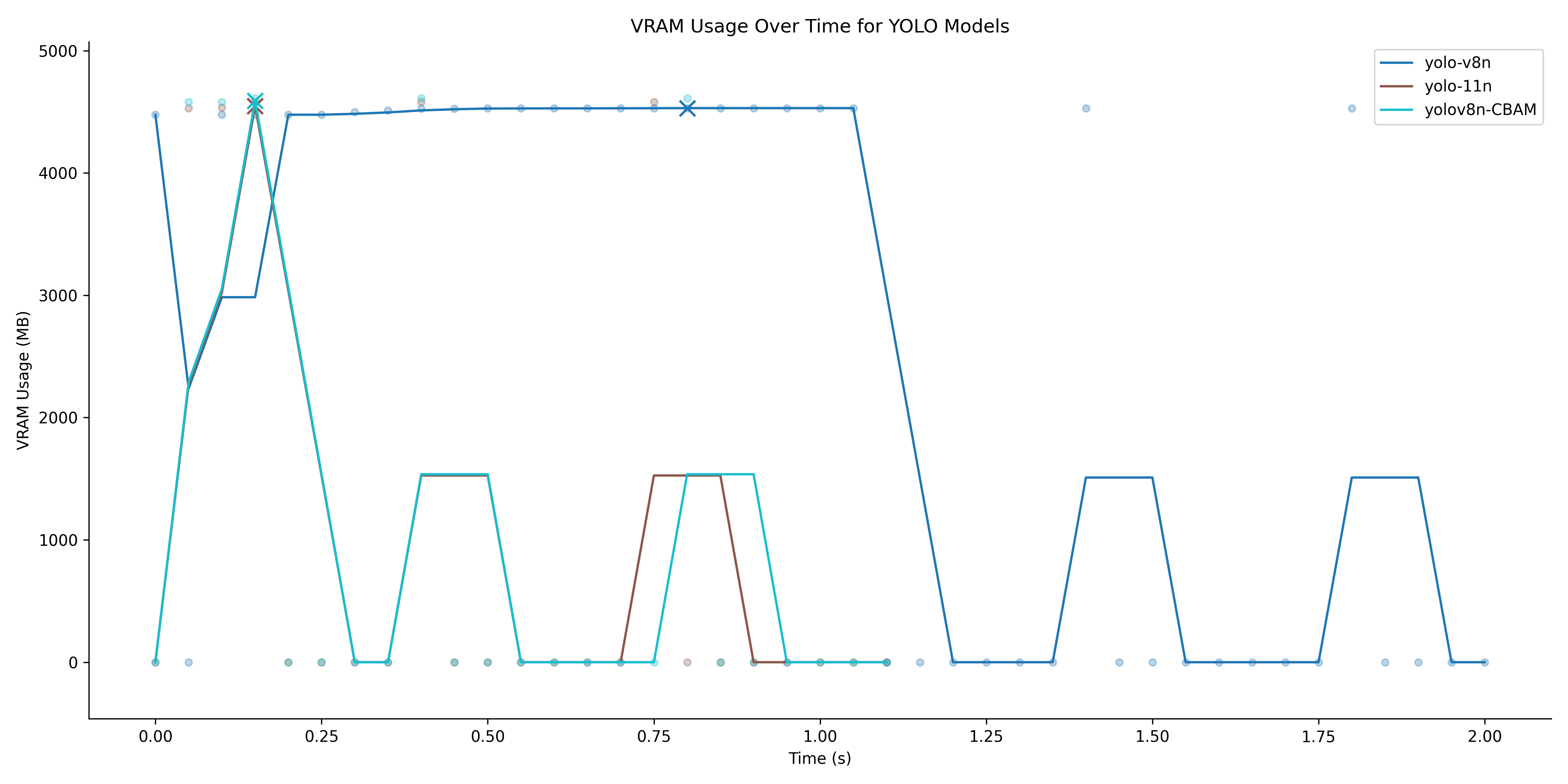}
  \caption{Object detection models VRAM usage under inference}
  \label{fig:object_detection_model_vram_usage}
\end{figure}

\begin{figure}[H]
  \centering
  \includegraphics[width=\columnwidth]{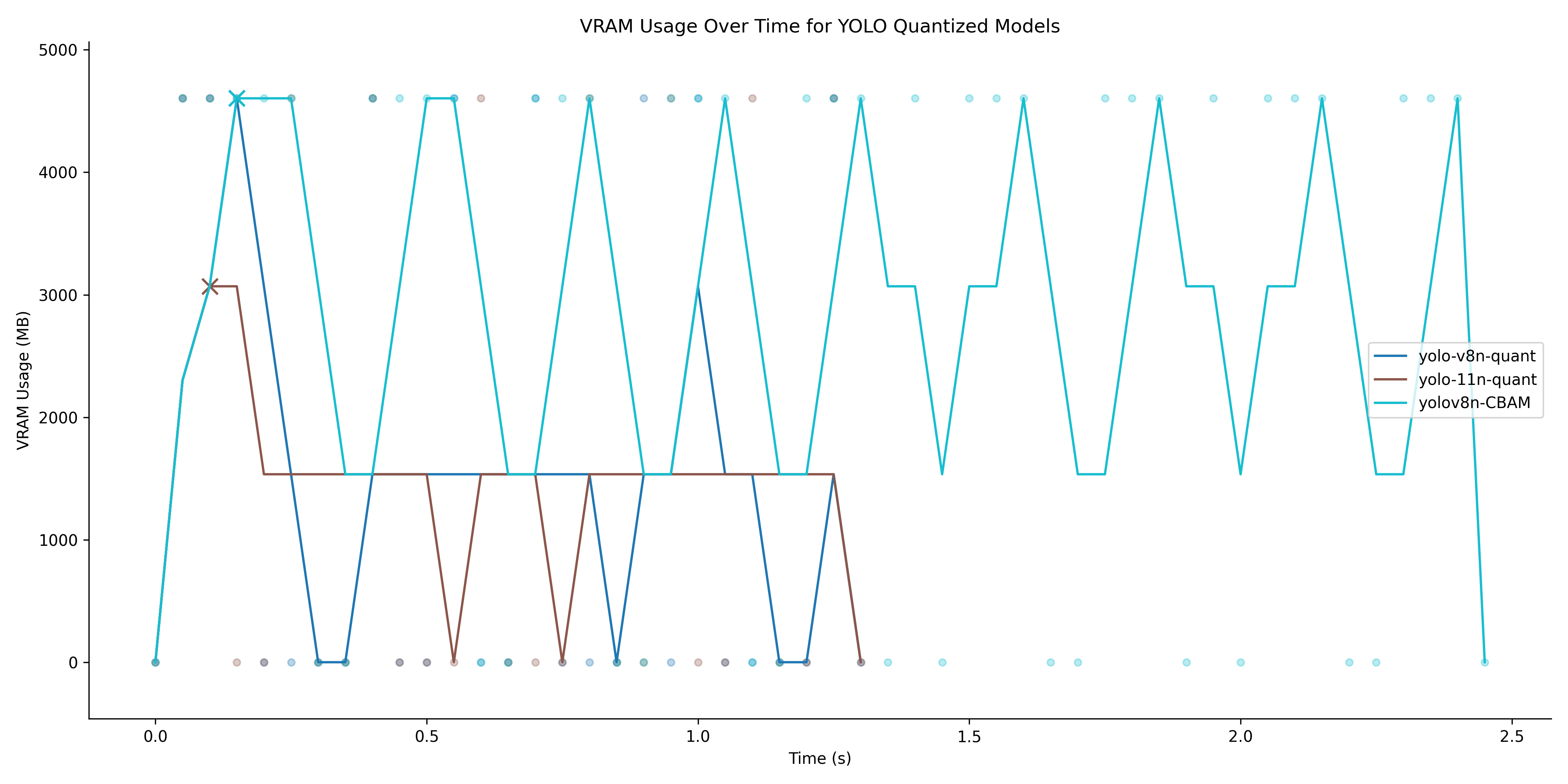}
  \caption{Object detection models VRAM usage after quantization under inference}
  \label{fig:object_detection_model_quantization}
\end{figure}

While all classification models, including MobileNet, exhibited low VRAM usage during inference time as shown in Figure \ref{fig:classification_model_vram_usage} and Figure \ref{fig:object_detection_model_vram_usage}. In contrast, the YOLO models consumed slightly more VRAM initially as shown in Figure \ref{fig:object_detection_model_quantization} but achieved faster inference time, which further improved significantly following quantization.

Notably, YOLOv11n emerged as the optimal architecture for edge deployment, attaining the best mean Average Precision (mAP = 0.77) with the smallest quantized footprint (2.8 MB) and the fastest speed, confirming its efficacy for resource-constrained applications with minimal carbon emissions. Our proposed model performed best in terms of precision, recall, F1-score, and mAP before quantization, and also had low VRAM usage with compared to YOLOv8n and YOLOv11n. However, its VRAM usage increased significantly after quantization. 

\begin{figure}[H]
  \centering
  \includegraphics[width=\columnwidth]{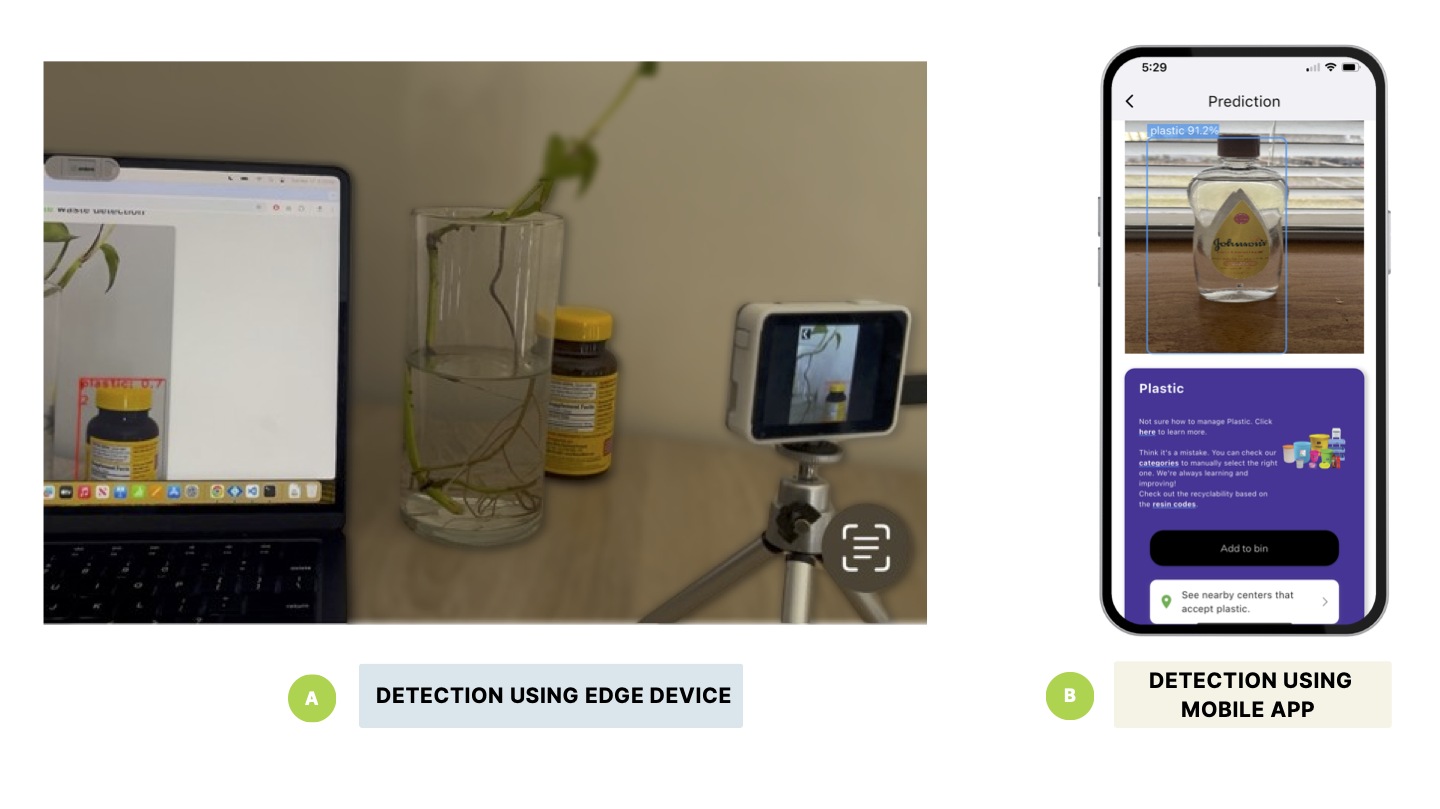}
  \caption{Real-world detection on edge and mobile device}
  \label{fig:real_world_deployment}
\end{figure}

The YOLOv8n-CBAM model has been successfully deployed to both the DWaste mobile application and a dedicated edge device for a real time waste object. An example of waste detection using both the app and edge device is shown in Figure \ref{fig:real_world_deployment}. 

\section{Conclusion}
This study successfully quantified the inherent trade-off between model accuracy and energy efficiency for deep learning-based waste sorting systems. Our results clearly demonstrate that while larger architectures (EfficientNetV2S/M, ResNet101/50) achieve superior classification accuracy, they demand greater computational resources and incur significantly higher carbon emissions. Conversely, lightweight object detection models, specifically YOLOv11n and YOLOv8n-CBAM, strike a crucial balance, offering strong real-time performance (mAP 77\% and 78\%) with minimal resource overhead. Furthermore, we confirmed that the application of model quantization dramatically aids deployment, consistently and substantially reducing VRAM usage and model size, thus directly lowering the energy required for inference on edge and mobile devices. Given these findings, the YOLOv8n-CBAM model, which achieved the best combination of accuracy, inference speed, and minimal resource usage, was selected and successfully embedded into the DWaste mobile app and a dedicated edge device for practical waste sorting implementation. Future work should focus on refining lightweight model accuracy with an expanded dataset, and conducting a longitudinal study to evaluate the long-term effectiveness and economic impact of the deployed app and the edge system.

\bibliographystyle{unsrt}
\bibliography{main}

\begin{thebibliography}{10}

\bibitem{pinos_why_2022}
Juan Pinos, John~N. Hahladakis, and Hong Chen.
\newblock Why is the generation of packaging waste from express deliveries a major problem?
\newblock {\em Science of The Total Environment}, 830:154759, July 2022.

\bibitem{unep_beyond_2024}
{UNEP}, editor.
\newblock {\em Beyond an age of waste: turning rubbish into a resource}.
\newblock Number 2024 in Global waste management outlook. UNEP, Nairobi, 2024.

\bibitem{sala-garrido_monetary_2023}
Ramon Sala-Garrido, Manuel Mocholi-Arce, Maria Molinos-Senante, and Alexandros Maziotis.
\newblock Monetary valuation of unsorted waste: {A} shadow price approach.
\newblock {\em Journal of Environmental Management}, 325:116668, January 2023.

\bibitem{turner_problems_2024}
Keanah Turner and Younsung Kim.
\newblock Problems of the {US} {Recycling} {Programs}: {What} {Experienced} {Recycling} {Program} {Managers} {Tell}.
\newblock {\em Sustainability}, 16(9):3539, April 2024.

\bibitem{runsewe_machine_2023}
T.~Runsewe, H.~Damgacioglu, L.~Perez, and N.~Celik.
\newblock Machine learning models for estimating contamination across different curbside collection strategies.
\newblock {\em Journal of Environmental Management}, 340:117855, August 2023.

\bibitem{lu_computer_2022}
Weisheng Lu and Junjie Chen.
\newblock Computer vision for solid waste sorting: {A} critical review of academic research.
\newblock {\em Waste Management}, 142:29--43, April 2022.

\bibitem{liu_computer_2025}
Xinru Liu, Zeinab Farshadfar, and Siavash~H. Khajavi.
\newblock Computer {Vision}-{Enabled} {Construction} {Waste} {Sorting}: {A} {Sensitivity} {Analysis}.
\newblock {\em Applied Sciences}, 15(19):10550, September 2025.

\bibitem{lahoti_multi-class_2024}
Jayanti Lahoti, Jathin Sn, M.~Vamshi Krishna, Mallika Prasad, Rajeshwari Bs, Namratha Mysore, and Jyothi~S. Nayak.
\newblock Multi-class waste segregation using computer vision and robotic arm.
\newblock {\em PeerJ Computer Science}, 10:e1957, May 2024.

\bibitem{abo-zahhad_real_2025}
Mohammed~M. Abo-Zahhad and Mohammed Abo-Zahhad.
\newblock Real time intelligent garbage monitoring and efficient collection using {Yolov8} and {Yolov5} deep learning models for environmental sustainability.
\newblock {\em Scientific Reports}, 15(1):16024, May 2025.

\bibitem{ahmad_intelligent_2025}
Gulzar Ahmad, Fizza~Muhammad Aleem, Tahir Alyas, Qaiser Abbas, Waqas Nawaz, Taher~M. Ghazal, Abdul Aziz, Saira Aleem, Nadia Tabassum, and Aidarus~Mohamed Ibrahim.
\newblock Intelligent waste sorting for urban sustainability using deep learning.
\newblock {\em Scientific Reports}, 15(1):27078, July 2025.

\bibitem{khai_tran_deep_2023}
Thien Khai~Tran, Kha Tu~Huynh, Dac-Nhuong Le, Muhammad Arif, and Hoa Minh~Dinh.
\newblock A {Deep} {Trash} {Classification} {Model} on {Raspberry} {Pi} 4.
\newblock {\em Intelligent Automation \& Soft Computing}, 35(2):2479--2491, 2023.

\bibitem{soni_mobilenet-based_2024}
Tanishq Soni, Deepali Gupta, and Mudita Uppal.
\newblock {MobileNet}-{Based} {Garbage} {Classification}: {Enhancing} {Recycling} with {Machine} {Learning}.
\newblock In {\em 2024 {International} {Conference} on {Intelligent} {Computing} and {Emerging} {Communication} {Technologies} ({ICEC})}, pages 1--4, Guntur, India, November 2024. IEEE.

\bibitem{eva_urankar_waste_2025}
{Eva Urankar}.
\newblock Waste {Detection} on {Mobile} {Devices}: {Model} {Performance} and {Efficiency} {Comparison}.
\newblock {\em International Journal of Science and Research Archive}, 15(1):722--731, April 2025.

\bibitem{gelar_systematic_2025}
Trisna Gelar, Sofy Fitriani, and Setiadi Rachmat.
\newblock A {Systematic} {Literature} {Review} of {YOLO} and {IoT} {Applications} in {Smart} {Waste} {Management}.
\newblock {\em Green Intelligent Systems and Applications}, 5(2):123--139, August 2025.

\bibitem{xu_improved_2024}
Pei Xu, Xiaonan Luo, and Ji~Li.
\newblock Improved {YOLOv8} {Underwater} {Object} {Detection} {Combined} with {CBAM}.
\newblock In {\em 2024 {International} {Symposium} on {Digital} {Home} ({ISDH})}, pages 43--48, Guilin, China, November 2024. IEEE.

\bibitem{kunwar_managing_2024}
Suman Kunwar.
\newblock Managing {Household} {Waste} {Through} {Transfer} {Learning}.
\newblock {\em Industrial and Domestic Waste Management}, 4(1):14--22, March 2024.

\bibitem{suman_kunwar_dwaste-data-v4-annotated_2025}
Suman Kunwar.
\newblock dwaste-data-v4-annotated, 2025.

\bibitem{kunwar_annotate-lab:_2024}
Suman Kunwar.
\newblock Annotate-{Lab}: {Simplifying} {Image} {Annotation}.
\newblock {\em Journal of Open Source Software}, 9(103):7210, November 2024.

\bibitem{he_imbalanced_2013}
Haibo He and Yunqian Ma, editors.
\newblock {\em Imbalanced {Learning}: {Foundations}, {Algorithms}, and {Applications}}.
\newblock Wiley, 1 edition, June 2013.

\bibitem{singh_review_2023}
Simrandeep Singh, Harbinder Singh, Gloria Bueno, Oscar Deniz, Sartajvir Singh, Himanshu Monga, P.N. Hrisheekesha, and Anibal Pedraza.
\newblock A review of image fusion: {Methods}, applications and performance metrics.
\newblock {\em Digital Signal Processing}, 137:104020, June 2023.

\bibitem{benoit_courty_mlco2/codecarbon:_2024}
mlco2/codecarbon: v2.4.1, May 2024.

\bibitem{francy_edge_2024}
Samer Francy and Raghubir Singh.
\newblock Edge {AI}: {Evaluation} of {Model} {Compression} {Techniques} for {Convolutional} {Neural} {Networks}, September 2024.
\newblock arXiv:2409.02134.

\end{thebibliography}

\end{document}